\documentclass[journal]{IEEEtran}
\usepackage{amsmath,graphicx}
\usepackage{cite}
\usepackage{color}
\usepackage{setspace}
\usepackage{subfig}
\usepackage{multirow}
\usepackage{hyperref}

\graphicspath{{./figs/}}

\newcommand{\ie}{\emph{i.e. }}

\newcommand{\etal}{\emph{et al. }}

\title{Knowledge-based Analysis for Mortality Prediction from CT Images}
\author{Hengtao Guo, Uwe Kruger,~\IEEEmembership{Senior~Member,~IEEE}, Ge Wang,~\IEEEmembership{Fellow,~IEEE}, Mannudeep~K.~Kalra, Pingkun~Yan*,~\IEEEmembership{Senior~Member,~IEEE}
\thanks{Asterisk indicates corresponding author.}%
\thanks{H. Guo, U. Kruger and G. Wang are with the Department of Biomedical Engineering, Rensselaer Polytechnic Institute, Troy, NY 12180, USA (e-mail: guoh9@rpi.edu, wangg6@rpi.edu, krugeu@rpi.edu)}%
\thanks{M. K. Kalra is with the Department of Radiology, Massachusetts General Hospital, Boston, MA 02114, USA (e-mail: mkalra@mgh.harvard.edu)}%
\thanks{*P. Yan is with the Department of Biomedical Engineering and the Center for Biotechnology and Interdisciplinary Studies, Rensselaer Polytechnic Institute, Troy, NY 12180, USA (e-mail: yanp2@rpi.edu)}%
\thanks{This work was supported by National Heart, Lung, and Blood Institute (NHLBI) of the National Institutes of Health (NIH) under award R56HL145172.}
\thanks{The source code of this work is available at \href{https://github.com/DIAL-RPI/KAMP-Net}{https://github.com/DIAL-RPI/KAMP-Net}.}
}

\begin{document}
\maketitle
%
\begin{abstract}

Low-Dose CT (LDCT) can significantly improve the accuracy of lung cancer diagnosis and thus reduce cancer deaths compared to chest X-ray. The lung cancer risk population is also at high risk of other deadly diseases, for instance, cardiovascular diseases. Therefore, predicting the all-cause mortality risks of this population is of great importance. This paper introduces a knowledge-based analytical method using deep convolutional neural network (CNN) for all-cause mortality prediction. The underlying approach combines structural image features extracted from CNNs, based on LDCT volume at different scales, and clinical knowledge obtained from quantitative measurements, to predict the mortality risk of lung cancer screening subjects. The proposed method is referred as Knowledge-based Analysis of Mortality Prediction Network (KAMP-Net). It constitutes a collaborative framework that utilizes both imaging features and anatomical information, instead of completely relying on automatic feature extraction. Our work demonstrates the feasibility of incorporating quantitative clinical measurements to assist CNNs in all-cause mortality prediction from chest LDCT images. The results of this study confirm that radiologist defined features can complement CNNs in performance improvement. The experiments demonstrate that KAMP-Net can achieve a superior performance when compared to other methods. Our code is available at \href{https://github.com/DIAL-RPI/KAMP-Net}{https://github.com/DIAL-RPI/KAMP-Net}.
\end{abstract}

\section{Introduction}
\label{sec:intro}

\IEEEPARstart{L}{ow}-Dose CT has proven to be effective for lung cancer screening. For example, the National Lung Screening Trial (NLST) observed a 20\% decrease in lung cancer related mortality in at-risk subjects (55 to 74 years, 30 pack-year cigarette-smoking history) \cite{NLST_2011}. The prevalence of lung cancer is highly correlated with CVDs and both are associated with significant morbidity and mortality~\cite{pope2011lung,omenn1996effects}. More precisely, both share several risk factors that are predominantly attributed to unhealthy dietary habits, obesity and tobacco use etc.
By analyzing the NLST data, Chiles \etal \cite{chiles_association_2015} showed that coronary artery calcification (CAC) is strongly associated with mortality. In a different study, the Dutch-Belgian Randomized Lung Cancer Screening Trial (NELSON), it was found that CAC can predict all-cause mortality and cardiovascular events on lung cancer screening LDCT \cite{jacobs_coronary_2012}. The work in \cite{digumarthy_multifactorial_2018} has also shown significant difference in CAC scores between the survivor and non-survivor groups, indicating that CAC influences the mortality risk of lung cancer patients. Moreover, other factors may also increase the mortality risk. For example, non-surviving NLST subjects tend to have higher fat attenuation and decreased muscle mass, comparing to the surviving ones and there is a strong difference in emphysema severity between survivors and non-survivors \cite{digumarthy_multifactorial_2018}.


Over the past few years, the application of deep learning, a subdomain of machine learning, has led to a series of breakthroughs producing a paradigm shift that resulted in numerous innovations in medicine, ranging from medical image processing, to computer-assisted diagnosis, to health record analysis. Deep learning has also been applied for automatic calcium scoring from chest LDCT images. For example, Cano-Espinosa \etal \cite{cano2018automated} proposed to use a convolutional neural network for Agatston score regression from non-contrast chest CT scans without segmenting CAC regions. Recently, de Vos. ~\etal ~\cite{de2019direct} proposed to (i) use one convolutional network for template image and input CT registration and (ii) use another network for direct coronary calcium regression. Lessmann \etal \cite{lessmann_automatic_2018} report that (i) deep neural networks can measure the size of CAC from LDCT and (ii) the use of different filters, during the reconstruction process, can influence the quantification results. Training such networks, however, requires manually labeling the area of calcification from images. This results in significant efforts and only a small number of images can be annotated. This may adversely affect the network performance. Moreover, CAC segmentation does reveal other imaging markers that may predict the mortality risk.

\begin{figure*}
	\centering
\includegraphics[width=.9\textwidth]{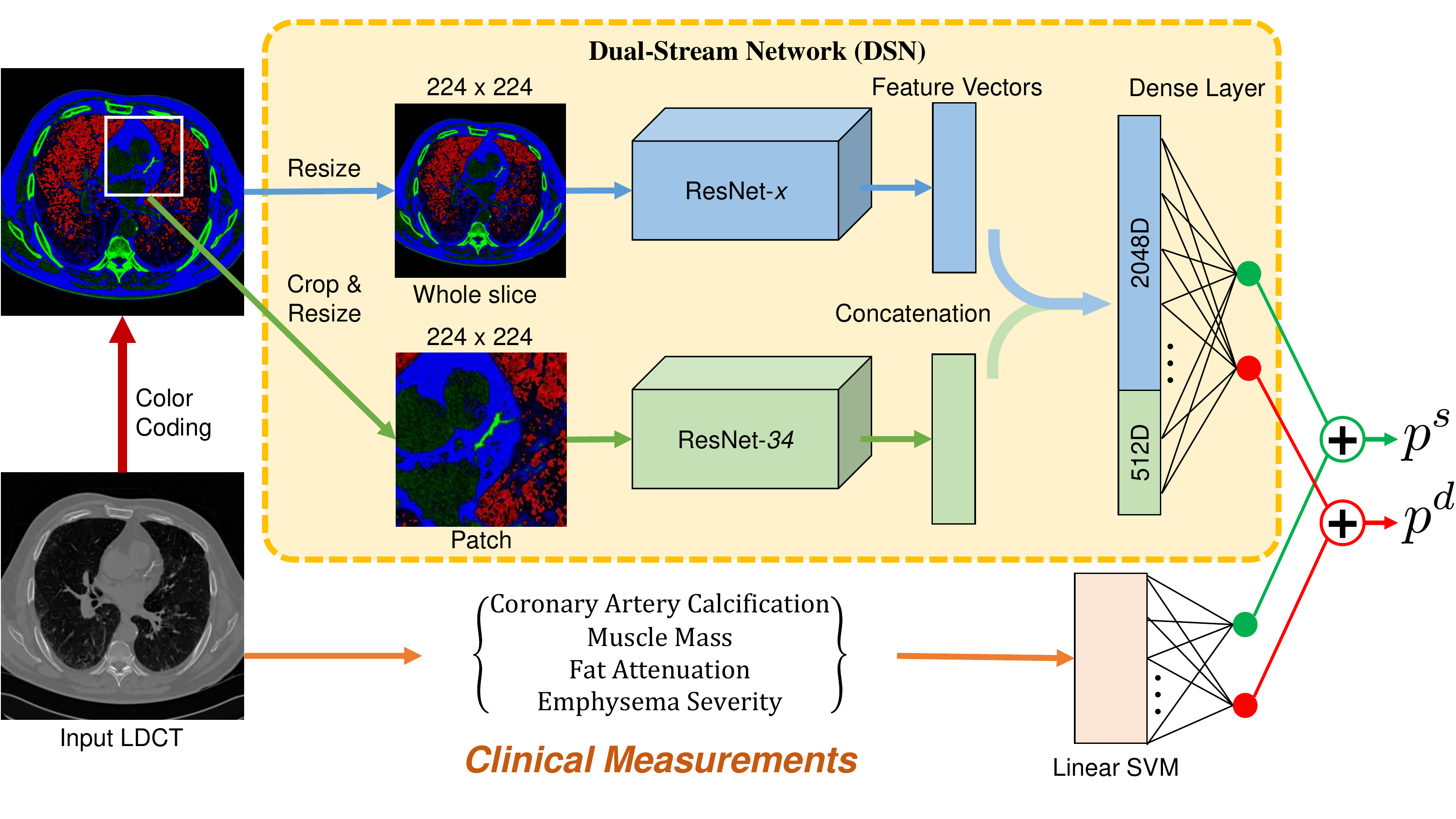}
\caption{Overview of the proposed KAMP-Net, which combines features extracted using clinical knowledge and features discovered by the deep learning DSN for improved mortality risk prediction.}
	\label{fig:kamp}
\end{figure*}

Recently, van~Velzen \etal \cite{van_velzen_direct_2018} introduced a convolutional autoencoder to extract image features for cardiovascular mortality prediction in a latent space. The features then serve as the input to a separate classifier, for example a neural network, a random forest classifier or a support vector machine, to compute a risk value. However, such a two-phase method may not be able to extract the most distinctive features associated with CVD. Moreover, traditional convolutional neural networks (CNNs) rely on directly extracted image features to perform image classification. This, however, omits clinical knowledge summarized by physicians through their diagnosis. Since various predefined imaging markers have been well recognized as indication of mortality risk, it is advisable to utilize this information for estimating this risk.


This paper hypothesizes that incorporating clinical knowledge into a deep learning based mortality risk prediction produces valuable complementary information which increases the prediction accuracy. To test the hypothesis, we introduce a novel method that combines extracted features from a CNN with clinical knowledge for predicting all-cause mortality risk of lung cancer patients from their LDCT images. More precisely, the method introduced here relies on a dual-stream network (DSN), which takes whole slices as well as cropped cardiac patches as the input for feature extraction. The multi-scale input has been demonstrated to have a positive impact on the CNN's performance~\cite{li2015visual}, as it contains both global image slice information and details of important local areas. The second component of the introduced method is incorporating clinical knowledge that is based on four clinical measurements, including CAC, muscle mass, fat attenuation, and emphysema. Inspired by the work of Fu \etal \cite{huazhu_multicontext_2018}, we employ a support vector machine (SVM) classifier to combine the clinical measurements to generate a combined mortality risk probability. The resultant method is referred to here as the knowledge-based analysis for mortality prediction (KAMP-Net).
The experimental results confirm that KAMP-Net predicts mortality more accurately when compared with other competitive networks. 
The contributions of this paper are summarized as follows.
\begin{enumerate}
  \item We utilize deep neural networks for predicting all-cause mortality risk of lung cancer patients by automatically discovering imaging features instead of measuring the extent of CAC as a surrogate index \cite{shemesh_coronary_2016,wolterink_automatic_2015}.
  \item We introduce a new gray-level image color-coding method to efficiently reuse the seminal deep CNN network structures.
  \item The DSN takes multi-scale image inputs, composed by LDCT slices and cardiac image patches, for both local and global feature extraction.
  \item Our results demonstrate that the DSN-extracted features, when combined with clinical knowledge from pre-defined imaging marker, can significantly improve the prediction performance.
\end{enumerate}


\section{Methods}
\label{sec:methods}

In this section, we present our proposed method for mortality risk prediction using LDCT images and related clinical measurements. 

\begin{figure}[tb]
\centering
\includegraphics[width=\columnwidth]{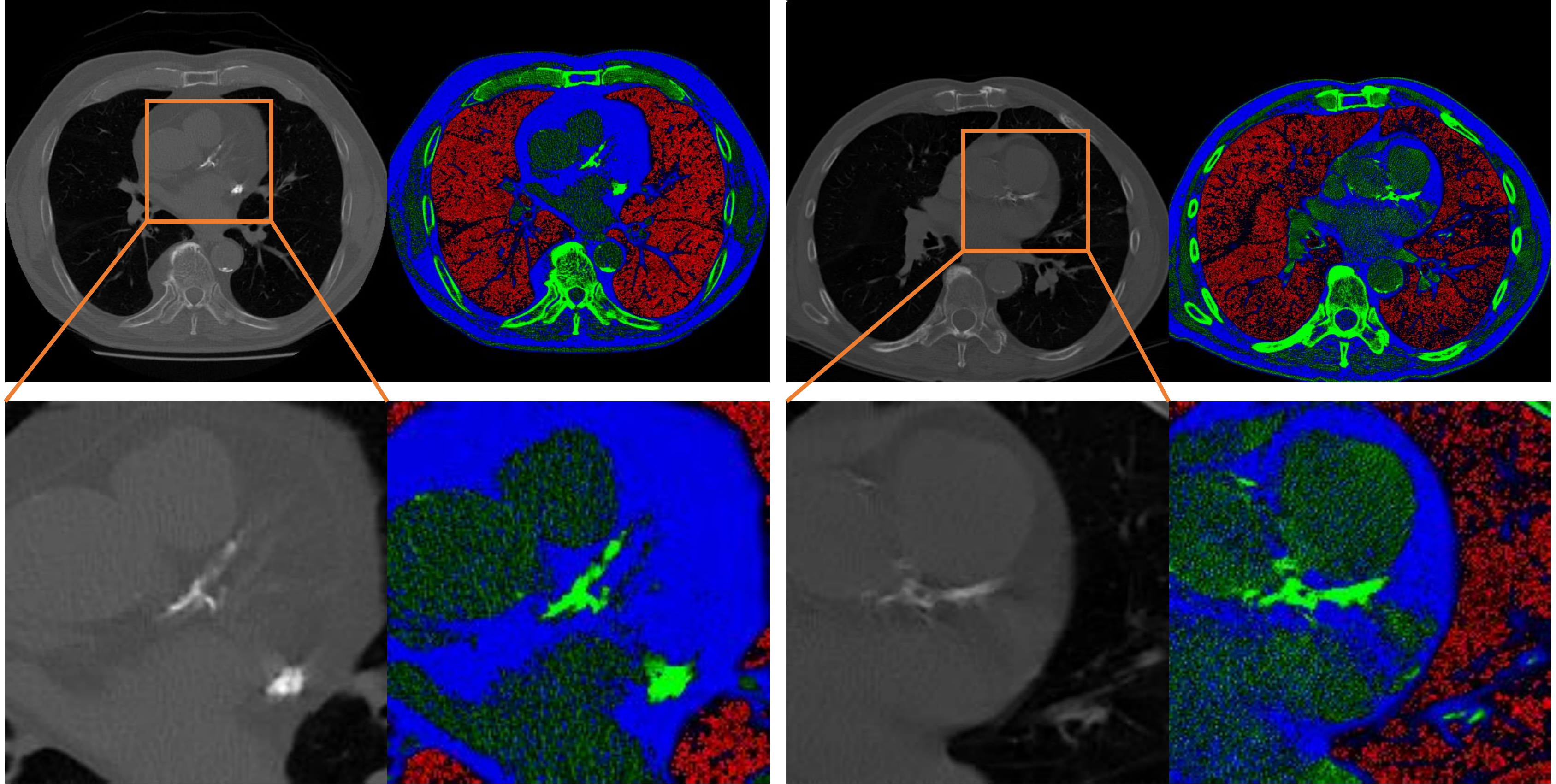}
\caption{Two examples of anatomical-information based multi-channel image coding. The second row is the magnification of the heart region in the first row. With the proposed coding scheme, the large intensity range of CT images can be divided into three smaller segments to highlight the important imaging features. Red areas mostly correspond to emphysema severity; blue areas represent fat attenuation concentrated regions; green areas contain mostly calcification or bone.}
\label{fig:colorcode}
\end{figure}

\subsection{Multi-Channel Image Coding}

LDCT images are 3D volume data containing information of internal structures such as organs, bones, blood vessels and soft tissue. The value of each voxel varies from -1000 Hounsfield units (HU) to around 2000 HU. Directly suppressing such a large value range into the typical range processed by deep CNN may result in information loss. To make full use of the anatomical information in CT images, we divide the range of CT image intensity values into three segments, according to the clinical expert knowledge on the intensity distribution of the tissues of interest. Namely, values below -900 HU are extracted and normalized to [0,255] as emphysema-concentrated interval to form the first channel. Similarly, voxels with values in the range of (-900,0] are assigned to the second channel representing fat-concentration intensity interval. CT numbers larger than 300 are typically from very strong calcification so we top off there and normalize all the values in (0,300] to form the third channel. For visualization purpose, the three channels are mapped to red, blue, and green channels of color image as shown in Fig.~\ref{fig:colorcode}. After separating different anatomical structures to separate channels, the intensity range of different tissue types throughout the CT slice become more balanced. For instance, the coronary artery calcification in the heart region appearing as bright green no longer suppressing other imaging components like fat or emphysema.

\subsection{Network Design and Implementation}

As shown in Fig.~\ref{fig:kamp}, the deep neural network consists of two streams, which is referred as dual stream networks (DSN). The upper stream extracts global image features from the input axial view image slice, which is manually chosen from LDCT scan. The lower stream takes one automatically selected region of interest (ROI) as input, which contains the most severe calcification in either left anterior descending (LAD), left circumflex (LCX) or left coronary artery (LCA). When there is no obvious calcification, the slices where the LAD is most visible were chosen. The automatic ROI detection is performed by a pre-trained cascaded detector \cite{viola2001rapid}. We implement both networks with 2D convolutions to guarantee a manageable computational burden. The lower stream supplements the upper stream with local detailed visual cues to emphasize the importance of those local regions. The lower stream supplements the upper stream with local detailed visual cues to emphasize the importance of those local regions.
The deep residual network (ResNet) \cite{he_deep_2016}, which is one of the top performing deep CNNs in various computer vision tasks, has been adopted as the backbone of DSN. By using only the convolutional layers of ResNet, image features can be extracted by ResNet-$x$, where $x$ denotes the depth of the network. At the end of the convolutional layers, 512 features are extracted by ResNet-18 and 34, and 2048 features are ResNet-50, 101 and 152, respectively. According to our previous work~\cite{pingkun_HyRiskNet_2018}, ResNet-34 achieves the best accuracy in the patch-input network, so we chose to use it as the lower stream's backbone architecture.


The proposed KAMP-Net was implemented in Python using the open source PyTorch library \cite{pytorch}. The training loss is defined as the cross-entropy between the prediction probability and ground-truth label as
\begin{equation}
  Loss=-\frac{1}{N}\sum_{i=1}^N(y_{i}\log_{}{(p_{i})}+(1-y_{i})\log_{}{(1-p_{i})})
\end{equation}
where N indicates the batch size, $y_{i}\in\{0,1\}$ is the label of groundtruth of the $i$th sample and $p_{i}$ is the network-derived probability for class $y_{i}$ after softmax. 
Training of the network is completed in two stages. The two streams of DSN are first trained separately in stage one and then combined for fine-tuning in stage two. 
%

In the first training stage, we implemented ResNet using the pre-defined structure provided by PyTorch \cite{pytorch}. Instead of generating probabilities for 1000 classes, the only difference between our network and the original ResNet is that the last fully connected (FC) layer outputs the classification probabilities of two categories: deceased or survived. Both patch-wise and slice-wise networks are trained from scratch using Adam optimizer~\cite{kingma2014adam} with initial learning rate of $1 \times 10^{-5}$, which then decays by 0.9 after every five epochs. While many DL-based medical image analysis papers report that networks pre-trained on ImageNet data can achieve better performance~\cite{shin2016deep}, we chose to train the network from scratch instead of using networks pre-trained on ImageNet data, because there exists large image appearance difference between natural images from ImageNet and the LDCT lung images.
Each sample in our dataset has been labeled either 0 (deceased) or 1 (survived) for training and validation.


In the second training stage, we remove the FC layers of the two sub-network streams pre-trained in stage one and combine the convolutional segments to form DSN. 
The output feature maps of the two sub-networks are concatenated and fed to a new FC layer, which generates two probabilities for survival and death prediction, respectively. The entire DSN with newly added FC layer is trained for another 200 epochs for fine-tuning with again the learning rate of of $1 \times 10^{-5}$. As the pre-trained slice-wise and patch-wise networks have already gained the ability to extract informative medical image features, the training of DSN would converge quickly.

\subsection{Integration of Deep Learning and Clinical Knowledge}

To further increase the accuracy of mortality prediction from LDCT images, we propose to combine clinical measurements with deep learning. Although CNNs are very powerful in extracting imaging markers, they lack of logical reasoning and high level intelligence of human experts, which makes it difficult for them to figure out connections between seemingly distant concepts. On the other hand, expert defined measurements from the images, including emphysema severity, muscle mass, fat attenuation and coronary artery calcification score, can be useful for this task as shown in the previous work ~\cite{digumarthy_multifactorial_2018}. CAC scores can be quantified in different ways~ \cite{agatston_quantification_1990, callister_coronary_1998} and automatic methods have been presented~\cite{gonzalez2016automated}. In our work, we utilize the CAC risk score, which was graded on a 4-point scale, to denote different severity. The CAC risk score is given by two radiologists, following a blinded and randomized manner. More detailed information about clinical measurements utilized in this work is available in the reference \cite{digumarthy_multifactorial_2018}. Those measurements contain high-level information and may not be readily grabbed by the CNNs. These knowledge based features can be complementary to what CNNs extract. We thus propose to combine the two groups of features to achieve more accurate prediction.

However, directly concatenating those measurements with the feature vectors from CNN could have only trivial effects on the prediction results. Since the CNN-extracted feature vector has much higher dimensionality (e.g. 512 for ResNet-34) than the clinical measurement (4 in this case), the latter will be overwhelmed after simple concatenation and contribute little to the risk prediction.
To balance the contributions of the two groups of features to the final output, we merge the two groups at a later stage after obtaining the initial probabilities. As shown in Fig.~\ref{fig:kamp}, a linear SVM classifier with the four clinical measurements as input is trained for mortality prediction. This SVM classifier will produce the probabilities of being deceased $p^d_{SVM}$ or survived $p^s_{SVM}$, which add up to 1. On the DSN side, a softmax activation function is used to generate the probability output. The two sets of probabilities are then combined to obtain the overall chance of survival as
\begin{equation}
  p^s = \alpha p^s_{DSN} + (1 - \alpha) p^s_{SVM},
\label{eqn:ratio}
\end{equation}
where $p^s$, $p^s_{DSN}$ and $p^s_{SVM}$ are the combined probability, DSN estimated probability and SVM estimated probability of survival, respectively. The contribution ratio $\alpha$ is a weighting parameter in the range of $[0,1]$. The probability of death $p^d$ can be computed as $1-p^s$.

\section{Experimental Results}
\label{sec:results}

This section presents experimental results of applying the KAMP-Net model for mortality risk prediction and provide detailed analysis and comparison of its performance.

\subsection{Materials}
\label{sec:materials}

All the study data used in this work are from the National Lung Screening Trial (NLST) \cite{chin_screening_2015}, which are managed by the National Cancer Institute Cancer Data Access System. In this large scale clinical trial, NLST compared LDCT with the chest radiography for lung cancer screening in more than 50,000 current or former smokers who met the various inclusion criteria. 
Our hypothesis of the study is that the analysis of LDCT images acquired for lung cancer screening can effectively predict the all-cause mortality of the subjects by combining the clinical knowledge and advanced deep learning techniques.
To efficiently investigate the effects of imaging features and clinical measurements, a balanced study is designed in our work. Following the same protocol used in \cite{digumarthy_multifactorial_2018}, 180 subjects were selected for the study, where the 90 survived and 90 deceased subjects are equally distributed in a variety of different cancer stages including no cancer. More precisely, each group consists of 49 subjects with stage I, 19 subjects with stage II, and 22 subjects with stage III lung cancers. The motivation is to rule out the influence of cancer stage and determine the effects of other factors, which may cause essential difference between the two groups.

The prediction is formulated as a binary classification problem by using the subject survival or decease status at the end of the follow-up period as the ground truth. The NLST trial uses lung cancer mortality as the primary endpoint of the study but also recorded all-cause mortality during the follow-up. The average follow-up period of the NLST trial is 6.5 years. More specifically, the average number of days of follow-up is $1660 \pm 488$ for the survivors, and the days to death for the deceased subjects are $894 \pm 542$. Each patient went through three LDCT lung cancer screening exams, of which the first LDCT scan of each patient is used in this study. The survival label is used as the ground truth for training and evaluating the prediction algorithms. 

The size of axial view slices in LDCT volume is $512 \times 512$ pixels. The number of slices per subject varies between 46 and 245. Three consecutive slices are extracted for each subject, which were manually chosen to be the slices in the CT volume for which the coronary artery is most visible. The use of three consecutive slices from a volume increases the number of slices from 180 to 540 images, \ie we have a significantly larger set for network training and validation.

Data augmentation has been shown to be an effective approach to improve the performance of deep CNNs~\cite{alex_alexnet_2012}. In this paper, data augmentation operations including random cropping and scaling are used for training the networks, which, theoretically, yields an infinite number of samples. The image patches in the size 161$\times$161 pixels are cropped from LDCT images using a pre-trained cascaded detector, which automatically locates a bounding-box over the heart region. Both the input slices and heart region patches are randomly cropped with the size ratio between 0.6 and 0.8. Please note this random ratio is the ratio of the original images. The cropping was conducted such that the aspect ratio is one, \ie length and height contain the same number of pixels. The cropped cardiac patches are resized to $224 \times 224$ pixels for network input. The aim of resizing the input images is to fit the design of original ResNet architecture.

Before applying to the ResNets, we conduct image normalization for gray-scale input and color-coded input separately, using their own means and standard deviations. As for the gray-scale images, we applied a single mean and a single standard deviation, which was computed from all the image samples. For the color-coded images, the mean and standard deviation are computed for each channel. For each channel, the normalization is performed by first subtracting its mean and then dividing the difference by its corresponding standard deviation. As a result, the pixel intensity distribution of the images has a mean of 0 and a standard deviation of 1 for each channel. In summary, the normalization for the gray scale slices and color-coded slices are performed separately, but in a consistent manner.


\subsection{Performance Evaluation}

\begin{figure}[tb]
	\centering
	\includegraphics[width=\columnwidth]{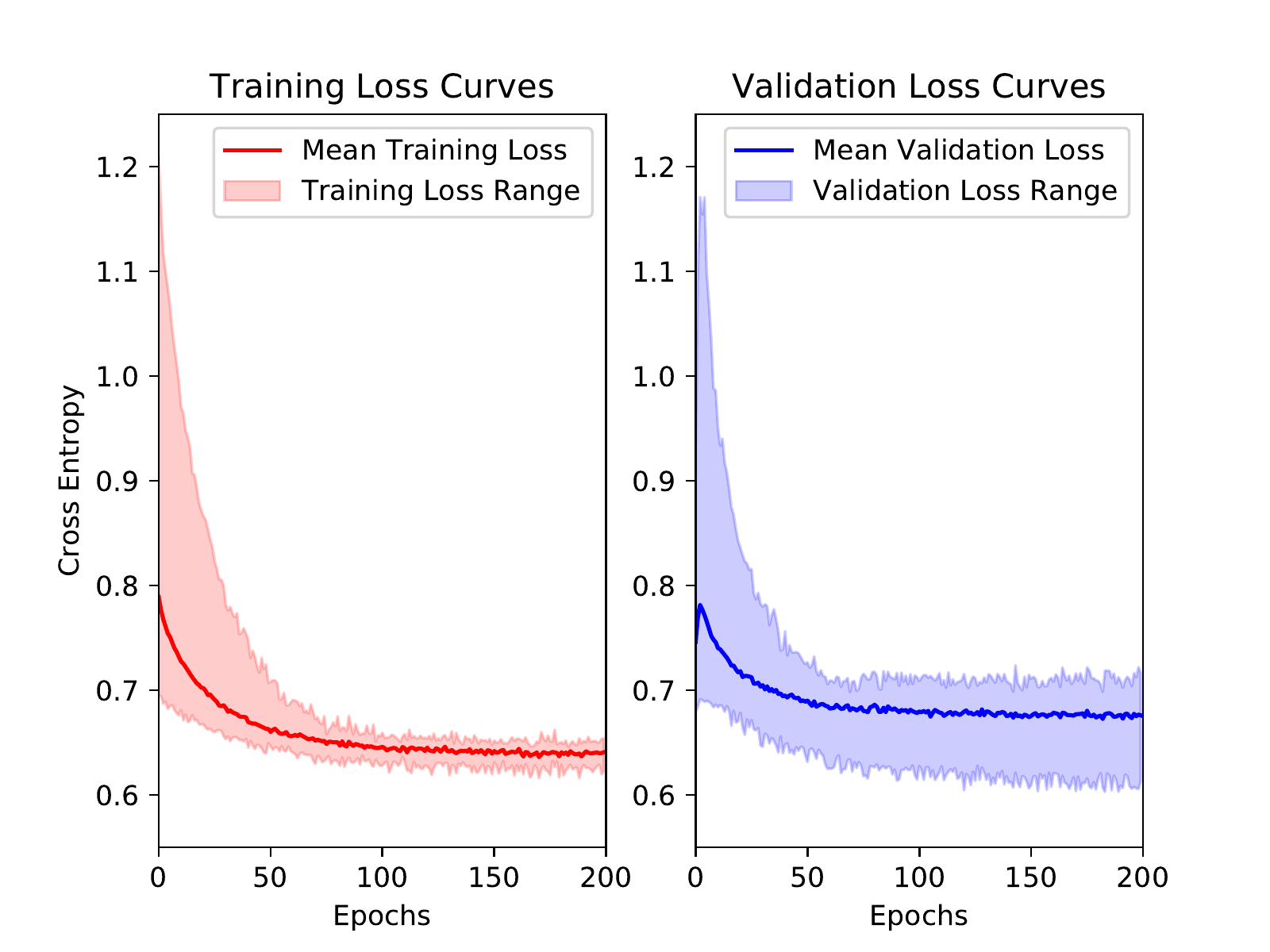}
	\caption{Training and validation loss curves of the ten-fold cross validation. The solid lines represent the mean loss curves over the ten folds and the shadow areas indicate the ranges.}
	\label{fig:loss_compose}
\end{figure}

Since the available dataset is relatively small, we applied a ten-fold cross validation scheme to our dataset for evaluating the performance of the proposed method and other comparative methods. We shuffle the list of subjects and divide them into 10 parts, where each part contains 18 subjects with 9 deceased and 9 survived. In each fold, one part is left out for testing. Among the remaining nine parts, one part is randomly chosen for validation and the other eight are for training. For each fold, the training is performed using the training set until the network performance is optimized on the validation set. Upon completion of this training process, the performance of the trained network is evaluated on the left-out testing set. The cross-validation continues until each part has been left out. In the testing phase, all three slices of each subject are used. We then compute the average probability and assign this average risk score to the subject. Since we aim to predict the ending points of subjects to be either ``survivor'' or ``nonsurvivor'' at the end of the follow-up period, receiver operating characteristic (ROC) curves are drawn to demonstrate the performance. Area under the curve (AUC) scores are used to compare the performances of different methods. When training the networks for each fold, the maximum number of epochs is set to be 200. Fig.~\ref{fig:loss_compose} shows the training and validation loss curves over a 200-epoch training of the ten-fold cross validation.


\begin{figure}[tb]
	\centering
	\includegraphics[width=\columnwidth]{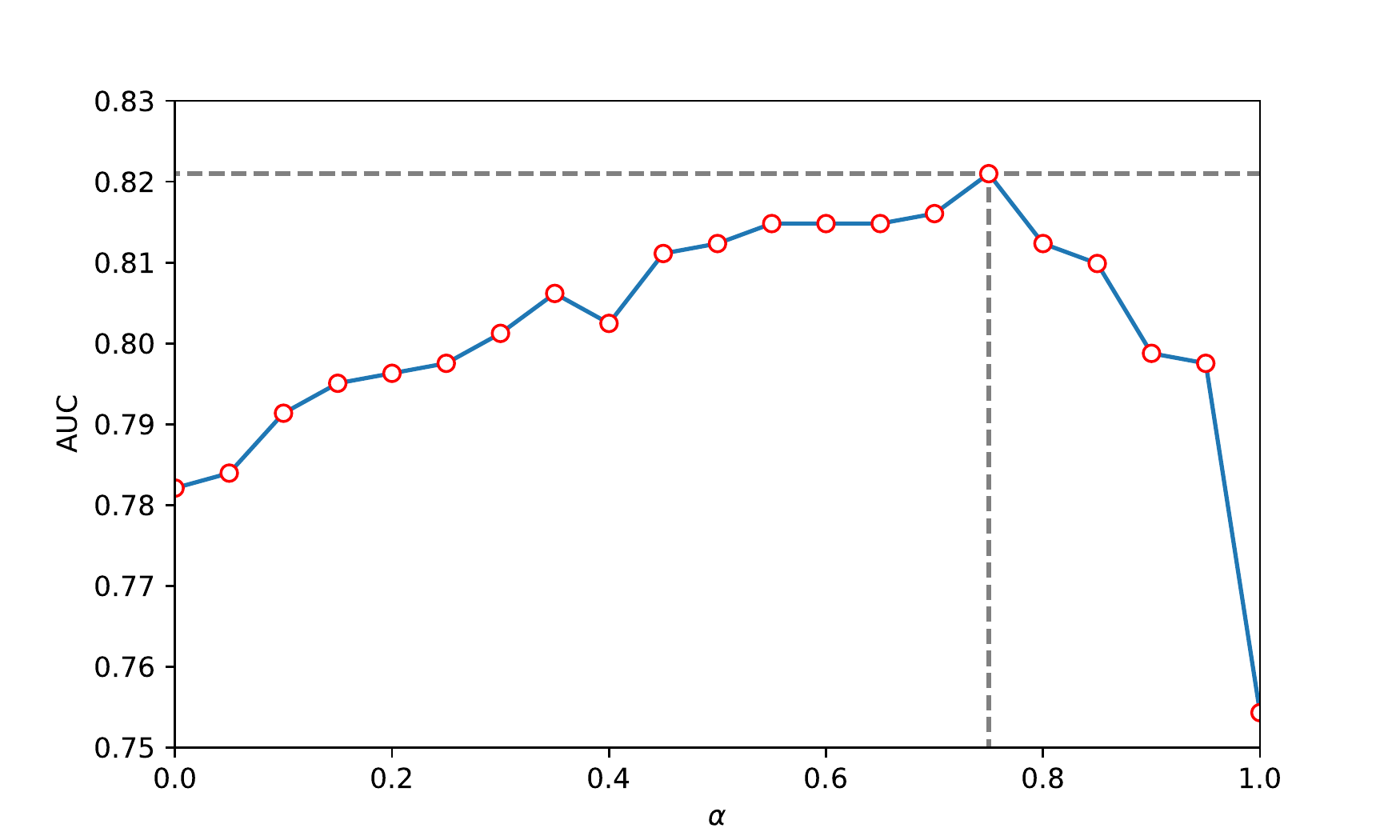}
	\caption{Effects of varying $\alpha$, the weight of probability votes from DSN in Eqn.~(\ref{eqn:ratio}), with increment of 0.05.}
	\label{fig:alpha_auc}
\end{figure}


To further evaluate the mutual influence between image-extracted features and the clinical information on the performance of the proposed KAMP-Net, we explore in terms of the DSN weight ratio $\alpha$ through 0 to 1, with an increment of 0.05. As shown in Fig.~\ref{fig:alpha_auc}, when the ratio $\alpha$ equals to 0.75, the curve arrives at its peak with the highest AUC score of 0.82 and the lowest standard deviation of 0.07. With ratio $\alpha$ increasing from 0 (pure SVM prediction votes from clinical measurements) to 1 (pure DSN prediction votes from LDCT images), the overall KAMP AUC score experienced a steady increase and then decrease. Such a tendency in this $\alpha$-AUC curve explicitly shows that there exists a delicate balance point where the votes from DSN and SVM can reach the best performance. At this balance point, the DL-based image features and the medical information from clinical measurements are collaborating with each other as well as complementing each other's missing clues on predicting one patient's status.

\begin{figure}[tb]
	\centering
	\includegraphics[width=\columnwidth]{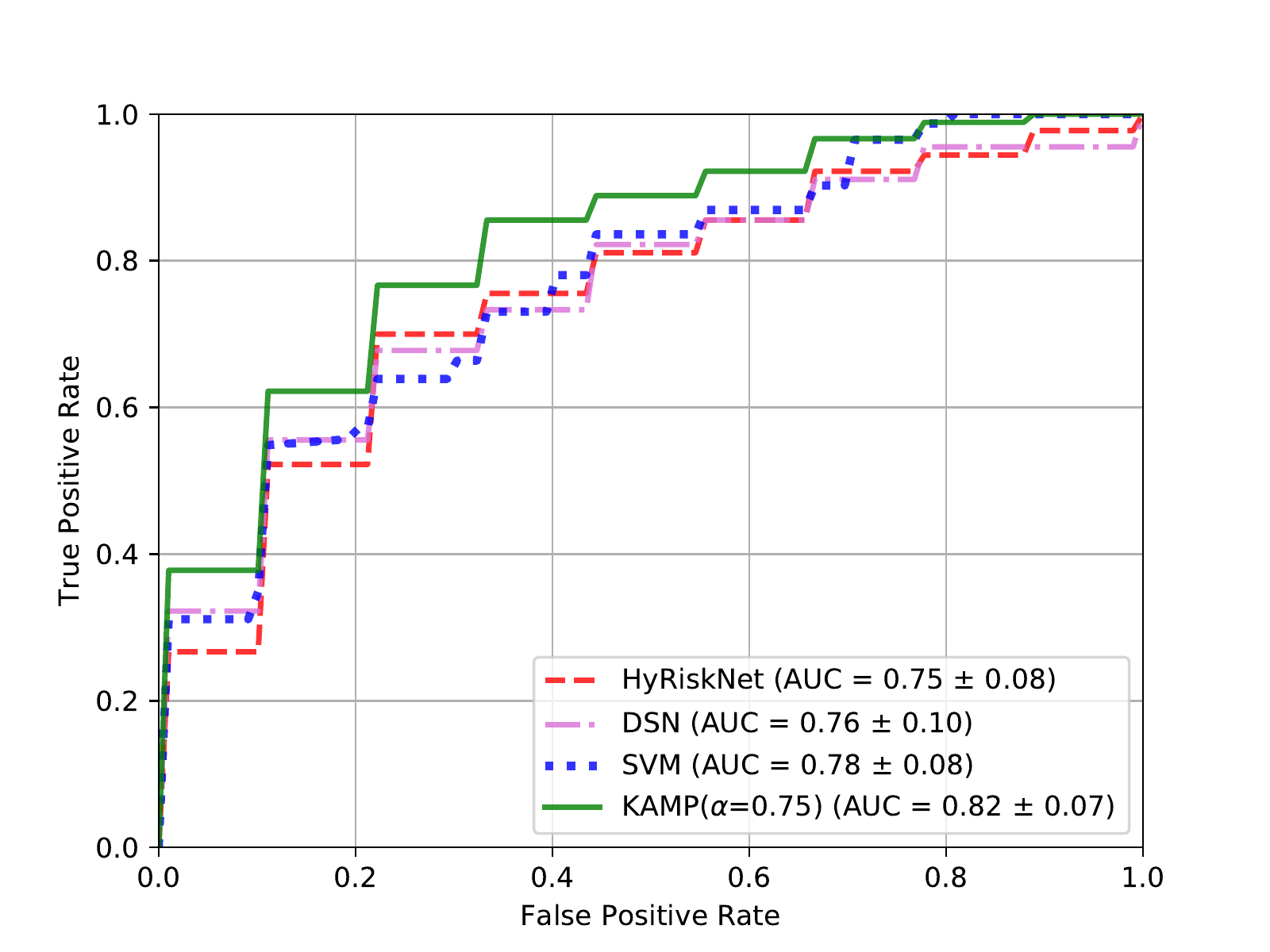}
	\caption{Ten-fold cross validation ROC curves and AUC values of HyRiskNet-34, DSN, SVM and the proposed KAMP-Net.}
	\label{fig:cross}
\end{figure}



The mean ROC curves over ten-fold cross validation of different methods are shown in Fig.~\ref{fig:cross}. The corresponding mean AUC scores and standard deviations are also provided. The SVM model is trained using the four clinical measurements on the same cross validation folds as the DL methods.
It can be seen that our proposed KAMP-Net achieves both the highest AUC score and the lowest standard deviation compared to other methods. 
Our previous work HyRiskNet~\cite{pingkun_HyRiskNet_2018} is included for comparison, which directly concatenates one additional CAC risk score with the high-dimensional deep CNN extracted feature vector.

We now compare the performance of KAMP-Net with that of its individual components, \ie the DSN and the SVM models. Fig.~\ref{fig:cross} allows comparing the performance of the three models graphically based on the estimated ROC curves. For KAMP-Net, we select $\alpha=0.75$. To qualitatively test whether the increase of the AUC value is statistically significant, we test the null hypotheses that AUC$_{\text{KAMP}}$ = AUC$_{\text{DSN}}$ and AUC$_{\text{KAMP}}$ = AUC$_{\text{SVM}}$ against the one sided alternative hypotheses AUC$_{\text{KAMP}}>$ AUC$_{\text{DSN}}$ and AUC$_{\text{KAMP}}>$ AUC$_{\text{SVM}}$, respectively. The two tests rely on three samples that store AUC values obtained from the previous 10-fold cross-validatory assessment; one sample stores the 10 AUC values for the KAMP-Net model, whilst the other two samples store the AUC values for the DSN and SVM models. This allows a pairwise comparison involving 10 pairs of values for testing AUC$_{\text{KAMP}}$ = AUC$_{\text{DSN}}$ as well as AUC$_{\text{KAMP}}$ = AUC$_{\text{SVM}}$. First, we confirmed that the two sample differences were drawn from normal distributions by applying the Anderson-Darling test and then applied a standard paired t-test~\cite{montgomery_ttest_2010}. In both cases, the null hypothesis was rejected and, therefore, concluded that the increase in the risk prediction accuracy by the KAMP-Net model is statistically significant.

It should be noted that even without using any clinical measurements, the current DSN has already outperformed the previous CNN based methods presented in~\cite{pingkun_HyRiskNet_2018}, which use only patch image information as input.
On the other hand, the performance of SVM shows that these four clinical measurements carry quantification information strongly associated with survival in our experiments.
However, it is only a limited set of measurements. When being complemented with deep CNN discovered features, the performance has become even better.

\begin{figure*}
	\centering
	\includegraphics[width=.8\textwidth]{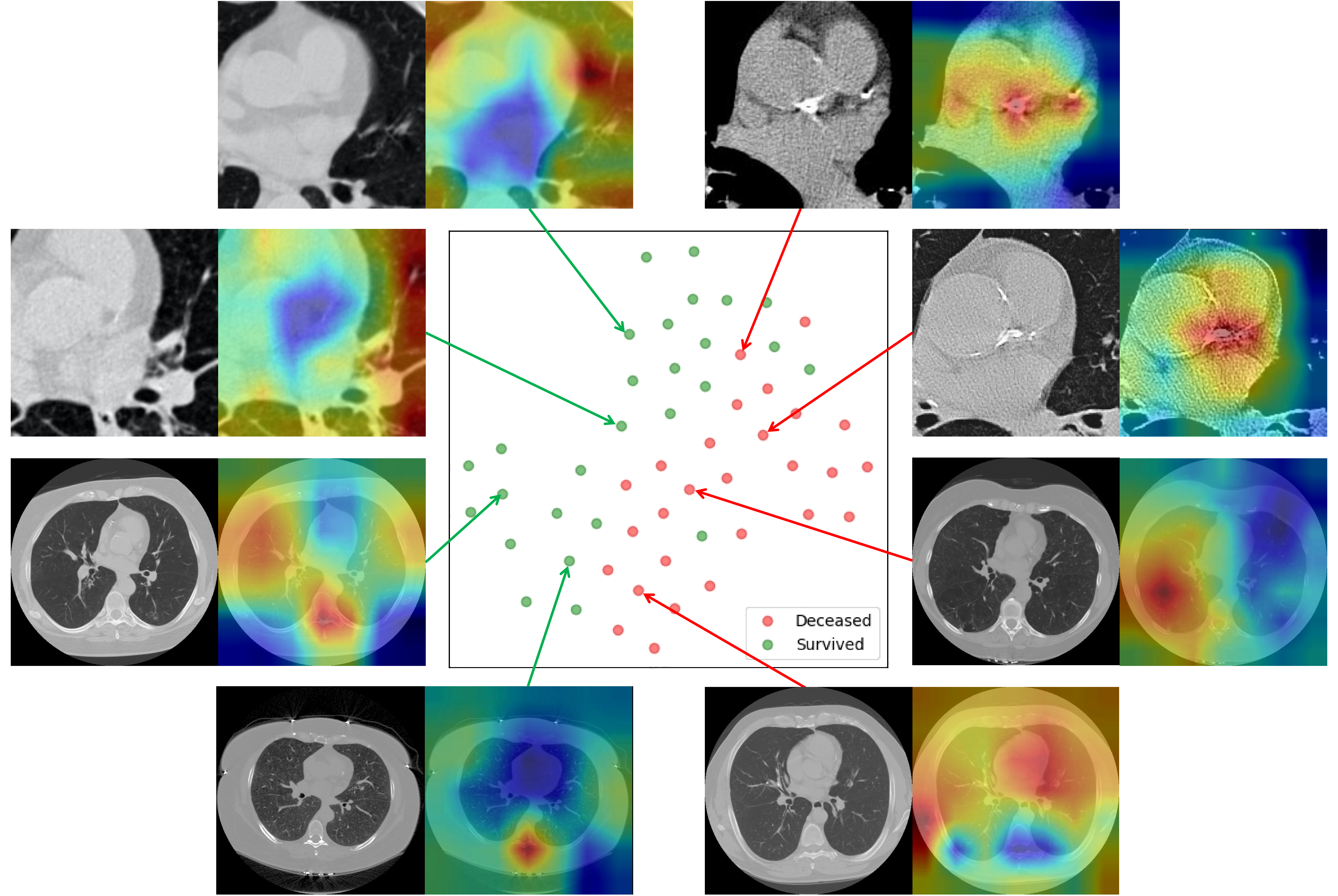}
	\caption{The scatter plot in the center is the t-SNE representation of combined feature vectors Each feature vector is the concatenation of one 2048D feature vector (from slice-input network stream) and one 512D feature vector (from patch-input network stream). Each green dot represents a survivor and a red dot denotes a non-survivor. The green dots and red dots are roughly separated from each other. Eight cases (four with cardiac patches and the other four with whole slices) are randomly selected to show the generated heatmaps.}
	\label{fig:crm}
\end{figure*}

\begin{table}[tbp]
	\centering
	\caption{Comparison of the performances of all ResNets and DSNs with color-coded and grey-scale LDCT image inputs, respectively. Slice-x and Patch-x indicate the ResNet-x architectures taking whole slice image input and cropped cardiac patch input, respectively. Two listed DSNs are composed of Slice-50 and Patch-34 networks trained on color-coding images and gray-scale images, respectively.}
	\label{tab:slice_cc}
	\begin{tabular}{l|c|c|c|c}
		\hline
		\multirow{2}{*}{Network} & \multicolumn{2}{c|}{Color-Coding} & \multicolumn{2}{c}{Grey-Scale} \\ \cline{2-5} 
		& AUC       & STD      & AUC       & STD      \\ \hline
		Slice-18 & 0.64 & 0.04 & 0.63 & 0.07     \\ 
		Slice-34 & 0.68 & 0.04 & 0.67 & 0.06     \\ 
		Slice-50 & \textbf{0.71}      & 0.06     & 0.68     & 0.05     \\ 
		Slice-101 & 0.68 & 0.06 & 0.65 & 0.08 \\ 
		Slice-152 & 0.66 & 0.07 & 0.64 & 0.07     \\ \hline 
		Patch-18 & 0.68 & 0.06 & 0.65 & 0.09  \\
		Patch-34 & \textbf{0.73}      & 0.07     & 0.68     & 0.07     \\ 
		Patch-50 & 0.69 & 0.04 & 0.66 & 0.04 \\
		Patch-101 & 0.66 & 0.03 & 0.65 & 0.05 \\
		Patch-152 & 0.64 & 0.04 & 0.64 & 0.07 \\
		\hline
		DSN & \textbf{0.76} & 0.10 & 0.70 & 0.08 \\\hline
	\end{tabular}
\end{table}

\subsection{Effectiveness of Color-Coding}

In this paper, we introduce the color-coding scheme to highlight the anatomical difference for more effective feature extraction. To evaluate the performance, we conducted experiments on all the ResNet network structures available in PyTorch using both the original LDCT image and the color-coded version as inputs, respectively. The experimental results are shown in Table.~\ref{tab:slice_cc}.
%
While the networks in Color-Coding group take the 3-channel pre-processed images as input, the networks in the other group just take the single-channel grey scale images as input. Such original images are obtained through directly suppressing the raw slices from LDCT 3D volume to the range [0,255] from a wide range of Hounsfield Units. The two groups of networks were trained on the same ten folds, with the same training strategy and parameters.

To statistically analyze the significance of color-coding, we applied a paired hypothesis test for the two groups of observations. Prior to that, we verified that the sample differences were drawn from a normal distribution by applying the well-known Anderson-Darling test. This allowed the use of the standard $t$-test~\cite{montgomery_ttest_2010} for the null hypothesis, which stated that the use of color-coded images does not affect the overall performance compared to gray-scale images, against the one-sided alternative hypothesis that color-coding increases the prediction accuracy when compared to gray-scale images.
The computed $t$-value of the slice-wise section maps to the rejection region and we, therefore, rejected the null hypothesis, which confirms that the use of color-coding led to a statistically significant improvement in the mortality prediction accuracy. This indicates that directly suppressing a whole slice from a large dynamic range to generate input for the networks may result in significant loss of information. Conversely, the introduced color-coding scheme alleviates this problem. In contrast, however, there is no significant difference between the color-coding group and the grey scale group when applying pre-processing to the patch-wise networks.
In summary of the results in Table~\ref{tab:slice_cc}, we select ResNet-50 and ResNet-34 as the backbone networks for the color-coded input slices and patches for DSN in KAMP-Net, respectively. Two DSNs, composed of Slice-50 and Patch-34 networks trained on color-coding input and gray-scale input respectively, achieve different performance. Such experimental results further indicate the superiority of applying color-coding scheme during the multi-scale analysis.

\begin{table}[tbp]
	\centering
	\caption{Comparison showing the effectiveness of DSN using pre-trained segments. All the following experiments are conducted on color-coded images.}
	\label{tab:dsn}
	\begin{tabular}{l|c|c}
	\hline
  Method   & AUC  & STD \\\hline
  DSN-scratch & 0.70 & 0.09 \\
  Slice-50  & 0.71 & 0.06 \\
  Patch-34  & 0.73 & 0.07 \\
  DSN     & \textbf{0.76} & 0.10 \\\hline
\end{tabular}
\end{table}

\subsection{Evaluation of Dual Stream Network}

We then evaluate the performance of DSN by comparing the network structures as well as training strategies. Table~\ref{tab:dsn} shows the ROC curves and also AUC values of DSN trained from scratch (SDN-scratch), Slice-50, Patch-34 and DSN.
It can be seen that DSN outperforms both Slice-50 and Patch-34 by combining them together and fine-tuning. This indicates that the slice- and patch-networks actually contain complementary information for each other, which leads to improved performance in the final mortality risk prediction. It is also interesting to see that DSN outperformed DSN-scratch by 8\% in terms of AUC score. That may be due to the difficulties in training the large concatenated network. The superior performance of our proposed DSN demonstrates the importance of having both well designed networks and good training strategy.


\subsection{Feature Visualization}

To help understand the features extracted by DSN, we compute the class activation map (CAM) by averaging the $512 \times 7 \times 7$ feature maps from the patch-wise network with the corresponding weights of the last FC layer as in~\cite{zhou2016learning}. We also used t-Distributed Stochastic Neighbor Embedding (t-SNE)~\cite{maaten_tsne_2008} to reduce the dimensionality of the feature maps to 2D for visualization. Fig.~\ref{fig:crm} shows the projection of validation samples from a randomly selected fold of the ten-fold cross validation scheme into 2D using t-SNE. From the point scattering shown in this figure, we can see that the positive and negative samples are roughly separated from each other, which indicates that DSN has the capability of extracting image features from LDCT images, which are strongly associated with the subject mortality.

Fig.~\ref{fig:crm} also includes several examples with CAMs superimposed on the gray scale images as heatmaps.
The closer to red in the heatmaps, the stronger activation there is in the original image, which indicates that information from that area contributes more to the final decision.
As it can be seen from Fig.~\ref{fig:crm}, the heatmaps for the deceased subjects predicted correctly by KAMP-Net tend to have strong activation over the coronary artery area in LDCT cropped cardiac areas, especially over the bright calcification region. This finding matches with the clinical literature that CAC is one of the major risk factors for mortality~\cite{chiles_association_2015}. For survived subjects, the heatmaps suggest that KAMP-Net looks more at surrounding lung tissue and muscles as suggested by our previous work in~\cite{digumarthy_multifactorial_2018}. For the heatmaps generated from image slices, survivors tend to have strong activation around the vertebral bone. It reflects the fact that subjects with higher bone density tends to be better health condition. In fact, two selected deceased subjects are both experiencing severe emphysema, and their generated heatmaps happen to highlight the emphysema region around the lungs.

\section{Discussions}

\begin{figure}[tb]
	\centering
	\includegraphics[width=\columnwidth]{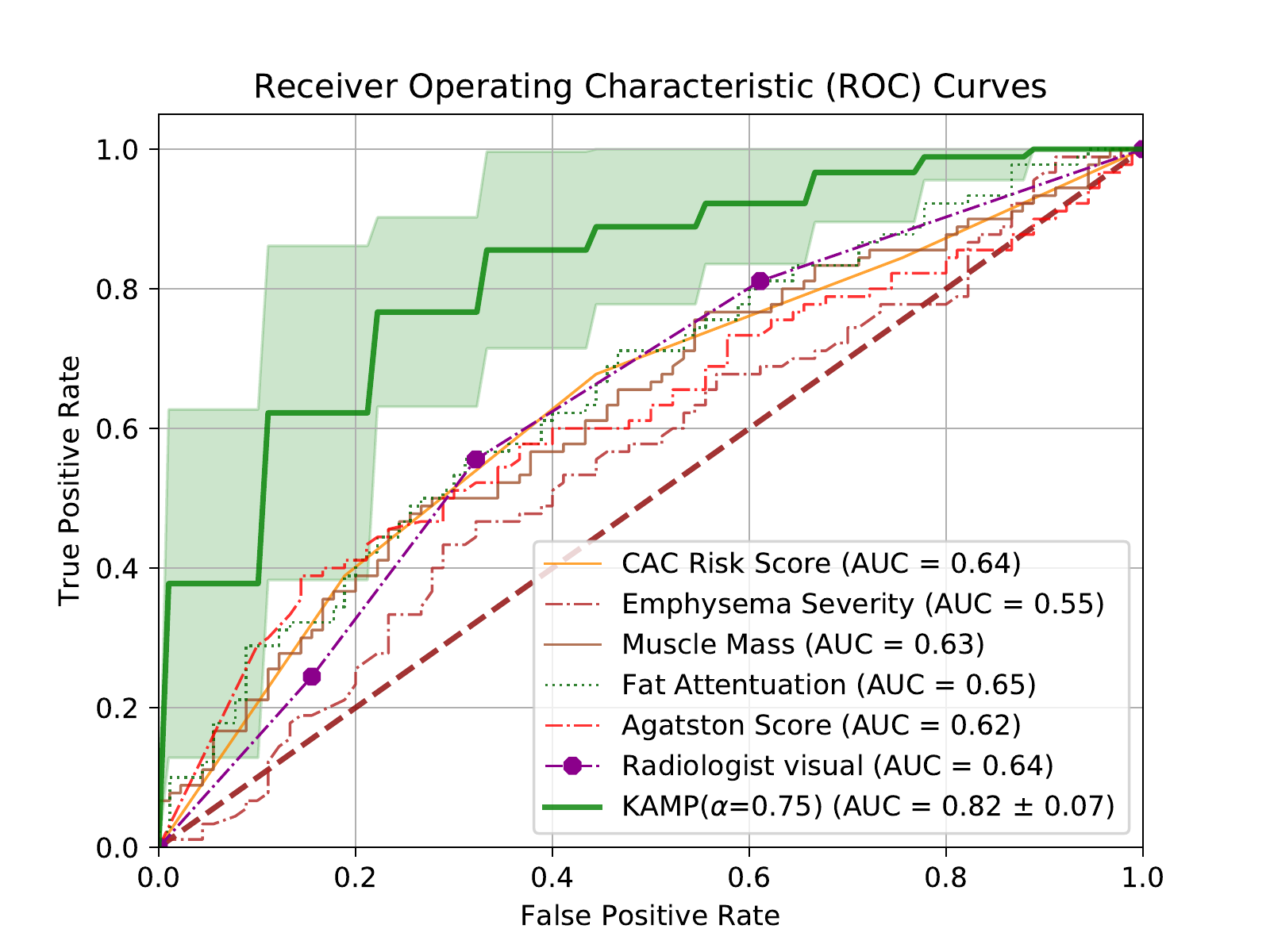}
	\caption{Performance comparison of various methods on all-cause mortality risk prediction.}
	\label{fig:compare}
\end{figure}


The developed KAMP-Net is then compared against several other clinically used scoring methods for further validation. The results are shown in Fig.~\ref{fig:compare}. 
It can be seen that the traditional semi-automatic methods, such as Agatston score~\cite{agatston_quantification_1990}, Agatston risk, muscle mass and fat attenuation perform similarly and the mean AUC values are in the range of [0.62, 0.65], which are slightly better than random guess. Emphysema severity itself alone cannot serve as a strong predictor (AUC = 0.55), which is consistent with the conclusions of a previous study~\cite{digumarthy_multifactorial_2018}.
It is interesting to see that the visual inspection of CAC by radiologists, with AUC of 0.64, outperforms the semi-automatic CAC scoring methods (Agatston score). This suggests that some information about the condition of cardiovascular vessels is not captured by those scoring methods, but has been taken into account by the radiologists.

The significant performance improvement comes from the proposed KAMP-Net as shown in Fig.~\ref{fig:compare}. The deep CNNs in DSN successfully extract and quantify features in cardiac patches and slices from chest LDCT images for all-cause mortality prediction, which couldn't be directly measured by radiologists. The proposed KAMP-Net (with $\alpha=0.75$) achieves the best performance with AUC of 0.82, which improves the prediction performance by 28.1\% over the visual inspection of radiologists.


\section{Conclusions}
\label{sec:conclusions}

In this paper, to accurately predict the all-cause mortality risk of a subject, we propose to combine multi-scale heterogeneous features. Those features are either automatically obtained from the images through training or manually defined by physicians based on their clinical knowledge. It has been shown that the patch-based and slice-based deep CNNs can complement each other in feature extraction for all-cause mortality prediction. Furthermore, incorporating the clinical measurements made by radiologists and summarized by a SVM model has yielded a significant performance improvement. This has led to the introduction of a novel method that combines the use of CNNs and a SVM models, which we have shown to produce a synergistic effect.

Our current study comes with the following limitations. 
\begin{enumerate}
  \item We manually choose the slices to cover the most significant CAC. In fact, we can improve the consistency of evaluation by automatically extracting the key slices from a 3D volume.
  \item The dataset used in our current work is of limited size. We will enlarge the dataset by including more subjects to evaluate the performance of the proposed method in our future work.
  \item The clinical measurements used in this study are manually acquired. It is, however, recommended to incorporate automatic scoring methods for future work.
  \item Although the color-coding pre-processing of LDCT images has shown to be beneficial, the current thresholds and channel arrangement were manually set, which could be performed automatically in our future work.
\end{enumerate}

\section{Acknowledgments}

The authors thank the National Cancer Institute for access to NCI's data collected by the National Lung Screening Trial. The statements contained herein are solely those of the authors and do not represent or imply concurrence or endorsement by NCI.
The authors would also like to thank NVIDIA Corporation for the donation of the Titan Xp GPU used for this research.

\bibliographystyle{IEEEtran}
\bibliography{refs}

\begin{thebibliography}{10}
\providecommand{\url}[1]{#1}
\csname url@samestyle\endcsname
\providecommand{\newblock}{\relax}
\providecommand{\bibinfo}[2]{#2}
\providecommand{\BIBentrySTDinterwordspacing}{\spaceskip=0pt\relax}
\providecommand{\BIBentryALTinterwordstretchfactor}{4}
\providecommand{\BIBentryALTinterwordspacing}{\spaceskip=\fontdimen2\font plus
\BIBentryALTinterwordstretchfactor\fontdimen3\font minus
  \fontdimen4\font\relax}
\providecommand{\BIBforeignlanguage}[2]{{%
\expandafter\ifx\csname l@#1\endcsname\relax
\typeout{** WARNING: IEEEtran.bst: No hyphenation pattern has been}%
\typeout{** loaded for the language `#1'. Using the pattern for}%
\typeout{** the default language instead.}%
\else
\language=\csname l@#1\endcsname
\fi
#2}}
\providecommand{\BIBdecl}{\relax}
\BIBdecl

\bibitem{NLST_2011}
{National Lung Screening Trial Research Team}, D.~R. Aberle, A.~M. Adams, C.~D.
  Berg, W.~C. Black, J.~D. Clapp, R.~M. Fagerstrom, I.~F. Gareen, C.~Gatsonis,
  P.~M. Marcus, and J.~D. Sicks, ``\BIBforeignlanguage{eng}{Reduced lung-cancer
  mortality with low-dose computed tomographic screening},''
  \emph{\BIBforeignlanguage{eng}{The New England Journal of Medicine}}, vol.
  365, no.~5, pp. 395--409, Aug. 2011.

\bibitem{pope2011lung}
C.~A. Pope~III, R.~T. Burnett, M.~C. Turner, A.~Cohen, D.~Krewski, M.~Jerrett,
  S.~M. Gapstur, and M.~J. Thun, ``Lung cancer and cardiovascular disease
  mortality associated with ambient air pollution and cigarette smoke: shape of
  the exposure--response relationships,'' \emph{Environmental health
  perspectives}, vol. 119, no.~11, pp. 1616--1621, 2011.

\bibitem{omenn1996effects}
G.~S. Omenn, G.~E. Goodman, M.~D. Thornquist, J.~Balmes, M.~R. Cullen,
  A.~Glass, J.~P. Keogh, F.~L. Meyskens~Jr, B.~Valanis, J.~H. Williams~Jr
  \emph{et~al.}, ``Effects of a combination of beta carotene and vitamin a on
  lung cancer and cardiovascular disease,'' \emph{New England journal of
  medicine}, vol. 334, no.~18, pp. 1150--1155, 1996.

\bibitem{chiles_association_2015}
C.~Chiles, F.~Duan, G.~W. Gladish, J.~G. Ravenel, S.~G. Baginski, B.~S. Snyder,
  S.~DeMello, S.~S. Desjardins, R.~F. Munden, and {NLST Study Team},
  ``\BIBforeignlanguage{eng}{Association of coronary artery calcification and
  mortality in the national lung screening trial: A comparison of three scoring
  methods},'' \emph{\BIBforeignlanguage{eng}{Radiology}}, vol. 276, no.~1, pp.
  82--90, Jul. 2015.

\bibitem{jacobs_coronary_2012}
P.~C. Jacobs, M.~J.~A. Gondrie, Y.~van~der Graaf, H.~J. de~Koning, I.~Isgum,
  B.~van Ginneken, and W.~P. T.~M. Mali, ``Coronary artery calcium can predict
  all-cause mortality and cardiovascular events on low-dose ct screening for
  lung cancer,'' \emph{American Journal of Roentgenology}, vol. 198, no.~3, pp.
  505--511, Mar. 2012.

\bibitem{digumarthy_multifactorial_2018}
\BIBentryALTinterwordspacing
S.~R. Digumarthy, R.~De~Man, R.~Canellas, A.~Otrakji, G.~Wang, and M.~K. Kalra,
  ``Multifactorial analysis of mortality in screening detected lung cancer,''
  \emph{Journal of Oncology}, vol. 2018, p.~7, 2018. [Online]. Available:
  \url{https://doi.org/10.1155/2018/1296246}
\BIBentrySTDinterwordspacing

\bibitem{cano2018automated}
C.~Cano-Espinosa, G.~Gonz{\'a}lez, G.~R. Washko, M.~Cazorla, and R.~S.~J.
  Est{\'e}par, ``Automated agatston score computation in non-ecg gated ct scans
  using deep learning,'' in \emph{Proceedings of SPIE--the International
  Society for Optical Engineering}, vol. 10574.\hskip 1em plus 0.5em minus
  0.4em\relax NIH Public Access, 2018.

\bibitem{de2019direct}
B.~D. de~Vos, J.~M. Wolterink, T.~Leiner, P.~A. de~Jong, N.~Lessmann, and
  I.~I{\v{s}}gum, ``Direct automatic coronary calcium scoring in cardiac and
  chest ct,'' \emph{IEEE transactions on medical imaging}, 2019.

\bibitem{lessmann_automatic_2018}
N.~Lessmann, B.~van Ginneken, M.~Zreik, P.~A. de~Jong, B.~D. de~Vos, M.~A.
  Viergever, and I.~Isgum, ``Automatic calcium scoring in low-dose chest {CT}
  using deep neural networks with dilated convolutions,'' \emph{IEEE
  Transactions on Medical Imaging}, vol.~37, no.~2, pp. 615--625, Feb. 2018.

\bibitem{van_velzen_direct_2018}
\BIBentryALTinterwordspacing
S.~G.~M. van Velzen, M.~Zreik, N.~Lessmann, M.~A. Viergever, P.~A. de~Jong,
  H.~M. Verkooijen, and I.~Išgum, ``Direct {Prediction} of {Cardiovascular}
  {Mortality} from {Low}-dose {Chest} {CT} using {Deep} {Learning},''
  \emph{arXiv:1810.02277 [cs]}, Oct. 2018, arXiv: 1810.02277. [Online].
  Available: \url{http://arxiv.org/abs/1810.02277}
\BIBentrySTDinterwordspacing

\bibitem{li2015visual}
G.~Li and Y.~Yu, ``Visual saliency based on multiscale deep features,'' in
  \emph{Proceedings of the IEEE conference on computer vision and pattern
  recognition}, 2015, pp. 5455--5463.

\bibitem{huazhu_multicontext_2018}
H.~Fu, S.~Xu, Yanwu abd~Lin, D.~W.~K. Wong, B.~Mani, M.~Mahesh, A.~Tin, and
  J.~Liu, ``Multi-context deep network for angle-closure glaucoma screening in
  anterior segment {OCT},'' in \emph{Medical Image Computing and Computer
  Assisted Intervention (MICCAI)}, Oct. 2018, pp. 356--363.

\bibitem{shemesh_coronary_2016}
J.~Shemesh, ``Coronary artery calcification in clinical practice: what we have
  learned and why should it routinely be reported on chest {CT}?'' \emph{Annals
  of Translational Medicine}, vol.~4, no.~8, Apr. 2016.

\bibitem{wolterink_automatic_2015}
J.~M. Wolterink, T.~Leiner, M.~A. Viergever, and I.~Išgum,
  ``\BIBforeignlanguage{en}{Automatic {Coronary} {Calcium} {Scoring} in
  {Cardiac} {CT} {Angiography} {Using} {Convolutional} {Neural} {Networks}},''
  in \emph{\BIBforeignlanguage{en}{Medical {Image} {Computing} and
  {Computer}-{Assisted} {Intervention} -- {MICCAI} 2015}}, ser. Lecture {Notes}
  in {Computer} {Science}.\hskip 1em plus 0.5em minus 0.4em\relax Springer,
  Cham, Oct. 2015, pp. 589--596.

\bibitem{viola2001rapid}
P.~Viola, M.~Jones \emph{et~al.}, ``Rapid object detection using a boosted
  cascade of simple features,'' \emph{CVPR (1)}, vol.~1, pp. 511--518, 2001.

\bibitem{he_deep_2016}
K.~He, X.~Zhang, S.~Ren, and J.~Sun, ``Deep {Residual} {Learning} for {Image}
  {Recognition},'' in \emph{{IEEE} {Conference} on {Computer} {Vision} and
  {Pattern} {Recognition} ({CVPR})}, Jun. 2016, pp. 770--778.

\bibitem{pingkun_HyRiskNet_2018}
P.~Yan, H.~Guo, G.~Wang, R.~De~Man, and M.~K. Kalra, ``Hybrid deep neural
  networks for all-cause mortality prediction from {LDCT} images,''
  \emph{arXiv:1810.08503 [cs.CV]}, Oct. 2018.

\bibitem{pytorch}
A.~Paszke, S.~Gross, S.~Chintala, G.~Chanan, E.~Yang, Z.~DeVito, Z.~Lin,
  A.~Desmaison, L.~Antiga, and A.~Lerer, ``Automatic differentiation in
  pytorch,'' in \emph{NIPS 2017 Workshop Autodiff}, 2017.

\bibitem{kingma2014adam}
D.~P. Kingma and J.~Ba, ``Adam: A method for stochastic optimization,''
  \emph{arXiv preprint arXiv:1412.6980}, 2014.

\bibitem{shin2016deep}
H.-C. Shin, H.~R. Roth, M.~Gao, L.~Lu, Z.~Xu, I.~Nogues, J.~Yao, D.~Mollura,
  and R.~M. Summers, ``Deep convolutional neural networks for computer-aided
  detection: Cnn architectures, dataset characteristics and transfer
  learning,'' \emph{IEEE transactions on medical imaging}, vol.~35, no.~5, pp.
  1285--1298, 2016.

\bibitem{agatston_quantification_1990}
A.~S. Agatston, W.~R. Janowitz, F.~J. Hildner, N.~R. Zusmer, M.~Viamonte, and
  R.~Detrano, ``Quantification of coronary artery calcium using ultrafast
  computed tomography,'' \emph{Journal of the American College of Cardiology},
  vol.~15, no.~4, pp. 827--832, Mar. 1990.

\bibitem{callister_coronary_1998}
T.~Q. Callister, B.~Cooil, S.~P. Raya, N.~J. Lippolis, D.~J. Russo, and
  P.~Raggi, ``Coronary artery disease: improved reproducibility of calcium
  scoring with an electron-beam {CT} volumetric method.'' \emph{Radiology},
  vol. 208, no.~3, pp. 807--814, Sep. 1998.

\bibitem{gonzalez2016automated}
G.~Gonz{\'a}lez, G.~R. Washko, and R.~S.~J. Est{\'e}par, ``Automated agatston
  score computation in a large dataset of non ecg-gated chest computed
  tomography,'' in \emph{2016 IEEE 13th International Symposium on Biomedical
  Imaging (ISBI)}.\hskip 1em plus 0.5em minus 0.4em\relax IEEE, 2016, pp.
  53--57.

\bibitem{chin_screening_2015}
J.~Chin, T.~Syrek~Jensen, L.~Ashby, J.~Hermansen, J.~D. Hutter, and P.~H.
  Conway, ``Screening for {Lung} {Cancer} with {Low}-{Dose} {CT} —
  {Translating} {Science} into {Medicare} {Coverage} {Policy},'' \emph{New
  England Journal of Medicine}, vol. 372, no.~22, pp. 2083--2085, May 2015.

\bibitem{alex_alexnet_2012}
A.~Krizhevsky, I.~Sutskever, and G.~E. Hinton, ``{ImageNet} classification with
  deep convolutional neural networks,'' in \emph{Advances in Neural Information
  Processing Systems (NIPS)}, 2012.

\bibitem{montgomery_ttest_2010}
D.~C. Montgomery and G.~C. Runger, \emph{Applied Statistics and Probability for
  Engineers, 5th Edition}.\hskip 1em plus 0.5em minus 0.4em\relax Hoboken, NJ:
  John Wiley \& Sons, 2010.

\bibitem{zhou2016learning}
B.~Zhou, A.~Khosla, A.~Lapedriza, A.~Oliva, and A.~Torralba, ``Learning deep
  features for discriminative localization,'' in \emph{Computer Vision and
  Pattern Recognition (CVPR)}, 2016, pp. 2921--2929.

\bibitem{maaten_tsne_2008}
L.~van~der Maaten and G.~Hinton, ``Visualizing data using {t-SNE},''
  \emph{Journal of Machine Learning Research}, vol.~9, no.~11, pp. 2579--2605,
  2008.

\end{thebibliography}

\end{document}